# Finding Black Cat in a Coal Cellar - Keyphrase Extraction & Keyphrase-Rubric Relationship Classification from Complex Assignments

Manikandan Ravikiran

mravikiran3@gatech.edu

*Abstract*—Diversity in content and open-ended questions are inherent in complex assignments across online graduate programs. The natural scale of these programs poses a variety of challenges across both peer and expert feedback including rogue reviews. While the identification of relevant content and associating it to predefined rubrics would simplify and improve the grading process, the research to date is still in a nascent stage. As such in this paper we aim to quantify the effectiveness of supervised and unsupervised approaches for the task for keyphrase extraction and generic/specific keyphrase-rubric relationship extraction. Through this study, we find that (i) unsupervised MultiPartiteRank produces the best result for keyphrase extraction (ii) supervised SVM classifier with BERT features that offer the best performance for both generic and specific keyphrase-rubric relationship classification. We finally present a comprehensive analysis and derive useful observations for those interested in these tasks for the future. . The source code is released in https://github.com/manikandan-ravikiran/cs6460-proj.

## 1 INTRODUCTION

Graduate programs with MOOC form of delivery (MOOC-Masters) are inherently dependent on peer and expert feedback (Joyner, 2017). While peer feedback focuses on incorporating pedagogical benefits, expert feedback renders an improved assessment. MOOC-Masters programs offer an inherent benefit of high scaling at lower costs. However, with the scale of these programs comes the problem of allocating multiple students to a single expert due to high monetary costs to students (Joyner, 2017) leading to delayed, inadequate feedback.

---

**Finding Black cat in a Coal Cellar** - This is because, we need to find phrases that are needed for purpose of scoring, which is a hard task due to its similarity with other contents in the assignments

Added to this, peer feedback is typically plagued by the problem of rogue reviews (Geigle, Zhai, and Ferguson, 2016) due to dishonesty, reprisal, competition or negligence of the peers (Kulkarni, Bernstein, and Klemmer, 2015).

Automatic grading systems are key in subduing the effects of the previously mentioned problems across both peer and expert feedback. However current automatic grading systems are typically confined to addressing a small part of one or more of these previously mentioned problems across various stages of the feedback process. More specifically, to date we have automatic scoring systems that estimate scores directly from textual essays, grading accuracy improvement tools that adjust scores based on aggregation, modeling, calibration and ranking strategies (Reily, Finnerty, and Terveen, 2009). To avoid bias and retaliation, we have tools that focus better on peer and expert allotments (Ardaiz-Villanueva et al., 2011). Finally, we also have review analysis tools that focus on understanding and enhancing the contents of a review comment.

Most of the existing works address the above problems positively yet, there are some lingering systematic issues including i) changes in reviewers rating with time and length of assignments leading to lack of effective response by the expert ii) variance between rating expected by the author to that of expert and the peer iii) random fluctuations of scores, etc. iv) lack of descriptive reviews owing to scale in both peer and expert feedback. Moreover, with complex assignments - Assignment with open-ended questions commonly seen in MOOC-masters programs with large and diverse content, the problems further exacerbate.

Considering these problems and the variety of complex assignments, we conjecture that the identification of important content needed for the feedback process would mitigate the previously mentioned issues. For example, extracting phrases that are needed for rubric alone would reduce time spent on the assignment and in turn lead to fruitful reviews. As such in this work, we focus on extraction and identification of relevant contents needed for the feedback process by organizing our work around following open research questions

- **RQ1: Keyphrase Extraction:** How effective supervised and unsupervised approaches for extracting important phrases required for peer feedback? The question is important because while supervised approaches are simple to build and unsupervised approaches are easy to scale across courses without any annotated data. *Through this study, we find that unsupervised ranking ap-* *proaches to be the ideal suit for keyphrase extraction from complex assignment with maximum F1 of 0.64.*

- **RQ2: Specific Keyphrase-Rubric Relationship Classification:** How do supervised, unsupervised and topic modeling approaches fair for specific keyphrase-rubric relationship classification? This question is critical because it relates the phrases directly to scoring rubrics. *We find supervised approaches to be more effective with maximum F1 of 0.48.*
- **RQ3: Generic Keyphrase-Rubric Relationship Classification:** How do traditional features compare against features from language models for generic keyphrase-rubric Relationship classification? The question is relevant because it tries to link the phrases to generic categories valid across multiple courses, thus concentrates on scaling. It specifically analyzes the effectiveness of language models and compares it with traditional approaches. *We find that both pretrained language models and traditional features like TF-IDF classifiers produce similar results with a former producing average of 0.06 F1 higher than the latter*

The rest of the paper is organized as follows. In section 2 we briefly describe the problem of keyphrase extraction and generic/specific keyphrase-rubric relationship classification. In section 3 we present literature related to research questions. Following this, in section 4 datasets, the annotation scheme and metrics used for evaluation are briefed. Algorithms used in this work are briefly explained in 5. In section 6 we present various experiments and results. Finally, in section 7, we conclude with a summary and possible implications for future work.

## 2 PROBLEM DEFINITION

In this section, we will present keyphrase extraction and generic/specific keyphrase-rubric relationship classification process.

**Keyphrase Extraction:** Consider Figure 1, given a complex assignment document as input keyphrase extraction focuses on the identification of phrases a.k.a combination of one or more sentences needed for peer feedback. Keyphrase extraction, depending on the algorithms used, is done in two different formulations. In the case of a supervised approach, keyphrase extraction is treated as a classification of phrase into keyphrase and non-keyphrase class. In the case of unsupervised approaches, keyphrase extraction approaches typically extract small parts of sentences (2-3 words), rather than the entire phrase. In such a case, we use fuzzy matching between the original phrases and extracted small parts

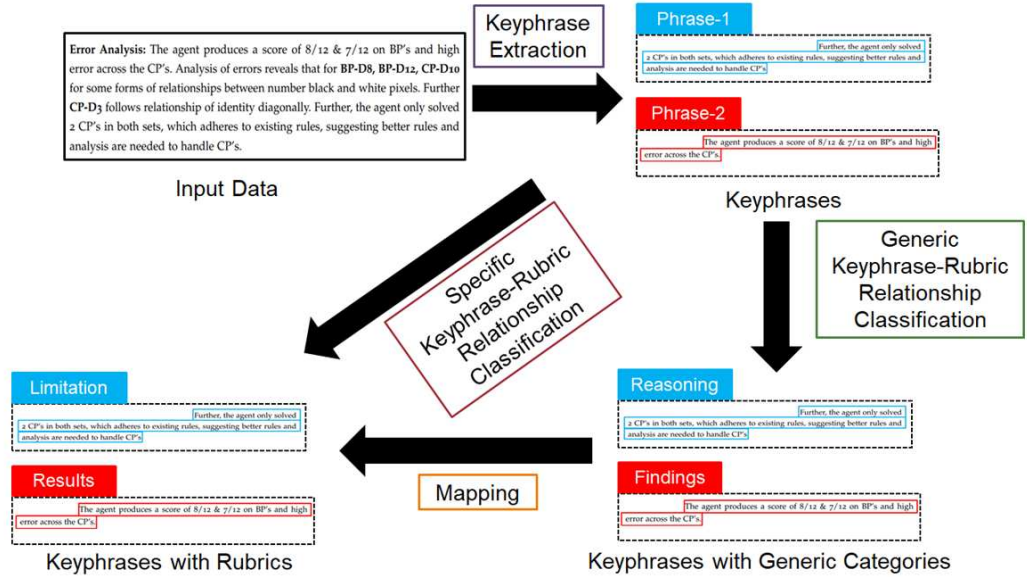

*Figure 1*—Overview of the automatic content extraction process from complex assignment with four major components. (1) Keyphrase Extraction, (2) Specific Keyphrase-Rubric relationship classification, (3) Generic Keyphrase-Rubric relationship classification and (4) Mapping of Generic classes to Rubrics.

to generate the final list of key phrases (See Figure 2).

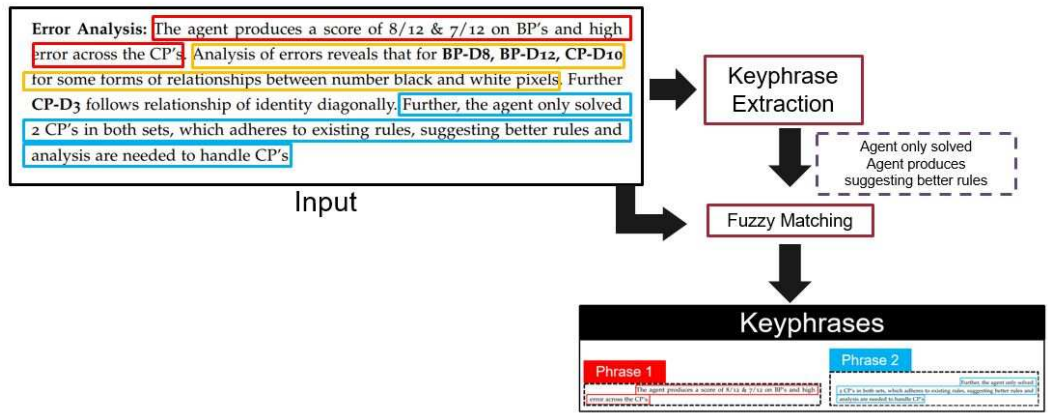

*Figure 2*—Keyphrase extraction process with Fuzzy Matching.

**Specific Keyphrase-Rubric Relationship Classification:** In this stage, Keyphrases Extracted by extraction algorithms are classified into a predefined set of scoring rubrics. For example in Figure 1a we can see that phrase-1 is mapped to scoring rubric of *results* and phrase-2 is mapped to scoring rubric of *limitation*. Thus we are automatically linking the contents to the scoring rubric, thus during peer feedback only the contents needed for rubric could be examined.

**Generic Keyphrase-Rubric Relationship Classification:** Previously, we directly classified each of the extracted keyphrases to set of predefined rubrics. However, depending upon courses the rubrics evolve or for a new course, the rubrics may be completely different. As a result, specific keyphrase-rubric relationship classification may need to re-adapted. Instead, we can create an intermediate mapping of the rubric to set of generic classes. Thus instead of directly classifying a phrase to a rubric, we can classify it to a generic class after which it could be mapped to the rubric. Consider figure 1b, here again, we extract the keyphrase, however instead of relating them directly to the scoring rubric, we classify it to a generic class. For example in figure 1b, phrase 1 is classified as *Finding* and phrase-2 is classified as *Reasoning*. Following this, we could have predefined mapping which states *Finding->Results* and *Reasoning->Limitation.* With generic keyphrase-rubric relationship classification, we can use similar generic classes across multiple courses.

In the next section, we shall present literature related to keyphrase extraction and keyphrase rubric relationship classification.

## 3 RELATED WORK

In this section, we present literature on various works that are closely related to the research questions so proposed in section 1.

### 3.1 Keyphrase Extraction

Keyphrase extraction focuses on extracting important phrases that signify a larger piece of content. A substantial body of works exists in both general and its application educational domain beginning with Zhang et al., (2016) which focused on the problem of automatically deriving keyphrases from tweets, extracting phrases from scientific documents (Boudin, (2018); Bhaskar, Nongmeikapam, and Bandyopadhyay, (2012)), Nguyen and Kan, (2007) and few more on educational applications (Badawy et al., (2018), Gollapalli, Li, and Yang, (2017); Al-Zaidy, Caragea, and Giles, (2019); Patel and Caragea, (2019); Wang et al., (2019); Zhang and Zhang, (2019)).

Then we have series of works on supervised approaches that treat keyphrase extraction as a supervised classification problem notable among them includes KEA (Witten et al., 1999) and WINGNUS (Nguyen and Luong, 2010), where both use information from text such as term frequency etc. for classification.

Unsupervised approaches, which have lately gained traction with much of the work being expressed as the ranking problem, notable and recent ones include (Bennani-Smires et al., 2018), which introduced an unsupervised key phrase extract approach from a single document using Embedrank. Similar works on ranking includes those TopicRank (Bougouin, Boudin, and Daille, 2013), TextRank (Mihalcea and Tarau, 2004), PositionRank (Florescu and Caragea, 2017), MultipartiteRank (Boudin, 2018). Finally, there are works of (Liu et al., 2009) that find exemplar terms by leveraging clustering techniques, which guarantees the document to be semantically covered by these exemplary terms.

In line with previous works we propose to extract key phrases from the educational text, however unlike them, we focus on complex assignments the key phrases are semi-formal.

### 3.2 Keyphrase-Rubric Relationship Classification

Keyphrase-rubric relationship classification is inherently a text classification problem. Text classification is a long-standing problem in educational technology, which has gone rapid increase with the advent of large scale MOOCs. The works vary according to their final intended goal itself, resulting in a diverse range of datasets, features, and algorithms.

The earliest works use a classification approach on clickstream datasets (Yang et al., 2015). Then there is works include that of Scott et al. (2015) which *analyzed three tools*. Then there are also sentiment related works, by (Ramesh et al., 2014) which included linguistic and behavioral features of MOOC discussion forums. Works on similar lines include (Liu et al., (2016); Tucker, Dickens, and Divinsky, (2014)).

On the parallel side, some works use posts and their metadata to detect confusion in the educational contents. Notable work by Akshay et al., (2015), *emphasizes the capacity of posts to improve content creation*. Additionally, there are works on post urgency classification (Omaima, Aditya, and Huzefa, 2018), speech act prediction (Jaime and Kyle, 2015). Finally, some works focus on using classification towards peer feedback, much of which focuses on scaffolding the review comments themselves (Xiong, Litman, and Schunn, 2010), (Nguyen, Xiong, and Litman, 2016), (Ramachandran, Gehringer, and Yadav, 2016),(Cho, 2008).

In the context of complex assignments, we have two major works of (Kuzi et al., 2019) and (Geigle, Zhai, and Ferguson, 2016) where both focus on automate grading of a medical case assessment using a supervised learning approach and introduce three general complementary types of feature representations.

Similar to the above works, we focus on the classification of educational text, however, there is three major difference in our work. Firstly, we focus on assignment text, rather than MOOC posts which is predominant in literature. Second, while our work on specific keyphrase-rubric classification is similar to the work (Kuzi et al., 2019) and (Geigle, Zhai, and Ferguson, 2016), we extend it to assignments of computer science graduate course and study impact of supervised, unsupervised and topic modeling approaches simultaneously. Finally, to the best of our knowledge, there exists no work on the task of generic keyphrase-rubric classification similar to our problem definition.

## 4 DATASET, ANNOTATION AND METHODS

In this section, we present the dataset and its annotation scheme, along with algorithms used across experiments.

### 4.1 Annotation Scheme

Annotation process used in this work is done in two phases where we first create annotation for generic keyphrase rubric relationship classification and in next phase we extend it to specific keyphrase extraction (section 4.1.1) and keyphrase extraction (section 4.1.2).

#### 4.1.1 *Annotation of generic keyphrase rubric relationship classification*

The annotation scheme used for generic key-phrase rubric relationship classification is as shown in Table 1. It consists of four different categories of important phrases that could be extracted and useful for grading complex assignments. The four different types of phrases include

- **Task** - Representing activity done by the author.
- **Findings** - Indicating the output of activity.
- **Reasons** - Showing reasons behind the findings.

- **Intuition** - Showing background on why the task was executed.

### 4.1.2 Extending Annotations

Table 1—Annotation Schema for generic keyphrase-rubric relationship classification

| Class | Description | Example |
|---|---|---|
| Task | Task indicates an activity explored by the student | We use using pixel based visual representations for images and develop an production system with series of rules |
| Findings | Findings indicates results of a Task | The agent only solved 2CP's in both sets, which adheres to existing rules, suggesting better rules and analysis are needed to handle CP's |
| Reason | Reason indicates the rationale behind the finding | Approximate similarity property can be seen in BP-E 9, but the result was erroneous |
| Intuition | Intuition represents the reason behind the Task | These problems satisfy simple relationships such as XOR, Overlay, Identity etc. |

In this phase, we link this to specific keyphrase rubric relationship classification and keyphrase extraction. The relative linking from Table 1 to annotations across keyphrase extraction, generic and specific keyphrase-rubric relationship extraction is as shown in Table 2.

Overall we annotate each sentence with three different labels. For example, from Table 2, the phrase *These problems satisfy simple relationships such as XOR, Overlay, Identity, etc.* will be considered as belonging to *keyphrase* class during keyphrase identification, *intuition* class during generic keyphrase-rubric relationship and will be related to *Agent Reasoning* class during specific keyphrase-rubric relationship classification.

Table 2—Mapping of coding schemes to classes in each tasks. We use same data across answering three different RQ's.

| Keyphrase Extraction | Generic Keyphrase-Rubric Relationship Classification | Specific Keyphrase-Rubric Relationship Classification | Example |
|---|---|---|---|
| Keyphrase | Task | Project Overview | We use using pixel based visual representations for images and develop an production system with series of rules |
| | Finding | Cognitive Connection | The agent only solved 2CP's in both sets, which adheres to existing rules, suggesting better rules and analysis are needed to handle CP's |
| | Reason | Relationship to KBAI class | Approximate similarity property can be seen in BP-E9, but the result was erroneous |
| | Intuition | Agent Reasoning | These problems satisfy simple relationships such as XOR, Overlay, Identity etc |
| Non-Keyphrase | **Other** | | The model was submitted on 11:45 GMT |

## 4.2 Dataset

The dataset in this work was developed using the four KBAI reports of Fall 2019. The overall dataset statistics are as shown in Table 3 below. The dataset consists of 791 phrases annotated according to Table 1. Out of this, we have 331 phrases that don't fit into any of the four classes. We annotated this as *Other*. The rest of 460 phrases are divided into train and test sets respectively[1]. Besides, the dataset consists of 2443 unique words with occurrences ranging from 1-800.

Further, the data is split into two folds used for overall cross-validation. Each of the phrases was subject to two rounds of annotation resulting in **Cohen's Kappa($\kappa$) of 0.77** for generic keyphrase-rubric relationship classification, showing that the resultant task is hard and may require more complex semantically relevant features.

Once upon creating dataset for generic keyphrase-rubric relationship classification, we created dataset for keyphrase extraction and keyphrase-rubric relationship classification through a simple mapping technique based on Table 2. For example, if we used *These problems satisfy simple relationships such as XOR, Overlay, Identity etc* as *Intuition*, then using Table 2 we added two more labels namely *keyphrase* and *agent reasoning* for keyphrase extraction and specific keyphrase-rubric relationship classification. Further, we can see that the dataset is imbalanced across the four classes with *Task* and *Finding* categories dominating the corpus and *Reason* being the least seen sentences. This behavior is because of the semi-formal nature of writing where the majority of sentences focus on inferences drawn from results.

*Table 3*—Dataset statistics used in this work.

| Splits | Train | | | | Test | | | | Other |
|---|---|---|---|---|---|---|---|---|---|
| | T | F | R | I | T | F | R | I | |
| **Fold-1** | 69 | 140 | 25 | 32 | 58 | 90 | 12 | 24 | |
| **Fold-2** | 71 | 140 | 22 | 43 | 56 | 90 | 23 | 15 | |
| **Total** | 276 | | | | 184 | | | | 331 |

## 4.3 Metrics

We use the following evaluation metrics in this work.

---

[1] Due to data privacy, the overall data that was used in this work is very limited. We plan to revisit this in the future.

- **Precision (P), Recall (R) and F1-Score (F1):** Precision is the ratio of correctly predicted positive observations to the total predicted positive observations. The recall is the ratio of correctly predicted positive observations to all observations in the actual class. F1 Score is the harmonic mean of Precision and Recall.
- **Cluster Purity (CP):** Homogeneity is computed by assigning each cluster to the class which is most frequent in the cluster, and then the accuracy of this assignment is measured by counting the number of correctly assigned phrases and dividing by N.

$$CP(\Omega, \mathbb{C}) = \frac{1}{N} \sum_k \max_j |\omega_k \cap c_j| \qquad (1)$$

where $\Omega = \{\omega_1, \omega_2, \ldots, \omega_K\}$ is the set of clusters and $\mathbb{C} = \{c_1, c_2, \ldots, c_J\}$ is the set of classes. We interpret $\omega_k$ as the set of phrases and $c_j$ as the set of input phrases which are classified.

- **Rand Index (RI):** An alternative to this *cluster purity* one can view clustering as a series of decisions, one for each of the $N(N-1)/2$ pairs of phrases in the collection. We want to assign two phrases to the same cluster if and only if they are similar. A true positive (TP) decision assigns two similar phrases to the same cluster, a true negative (TN) decision assigns two dissimilar phrases to different clusters. There are two types of errors we can commit. An (FP) decision assigns two dissimilar phrases to the same cluster. An (FN) decision assigns two similar phrases to different clusters. The Rand index (RI) measures the percentage of correct decisions. That is, it is simply accuracy calculated as

$$RI = \frac{TP + TN}{TP + FP + FN + TN} \qquad (2)$$

In this section, we shall present the feature extraction and supervised/unsupervised algorithms used for classification.

## 5 ALGORITHMS

In this section, we shall present various algorithms in brief used in section 6. A detailed presentation of the algorithm is beyond the scope of the current paper and we invite the readers to look at the respective original works. Overall we have feature extractors (see section 5.1) that is used to convert text to features, classification algorithms (see section 5.2), clustering algorithms (see section 5.4) and keyphrase extraction approaches (see section 5.5).

## 5.1 Feature Extraction

In this work, we use three different approaches for feature extraction namely BERT, TF-IDF and Latent Dirichlet Allocation(Blei, Ng, and Jordan, 2003) coupled with the former two.

1. **BERT**: BERT (Devlin et al., 2019) is a bidirectional encoder based on transformer stack. In this work, we use a bert-base-uncase model with a fully connected layer of 768 accounts and 12 attention heads. BERT was originally trained on Wikipedia and Book Corpus, a dataset containing +10,000 books of different genres. In our work, we use BERT works like a transformer encoder stack, by taking a sequence of words as input which keeps flowing up the stack from one encoder to the next, while new sequences are coming in. The final output for each sequence is a vector of size 728. We will use such vectors for our phrase classification problem without fine-tuning.

2. **XLNet:** XLNet (Yang et al., 2019) is a large bidirectional transformer that uses improved training methodology, larger data (130GB) and more computational power (512 TPU's) to achieve better than BERT prediction metrics on 20 language tasks. To improve the training, XLNet introduces permutation language modeling, where all tokens are predicted but in random order. This contrasts with BERT's masked language model where only the masked (15%) tokens are predicted. This helps the model to learn bidirectional relationships and therefore better handles dependencies and relations between words. Unlike BERT XLNET uses Transformer XL as the base architecture. .

3. **RoBERTa**: RoBERTa (Liu et al., 2019) from Facebook, robustly optimized BERT approach (RoBERTa), is a retraining of BERT with improved training methodology, 1000% more data and compute power. RoBERTA again uses 160GB of data for pretraining and introduces larger batch-training, dynamic masking so that the masked token changes during the training epochs. As a result, RoBERTa outperforms both BERT and XLNet on GLUE benchmark results.

4. **XLM:** Though BERT was trained on over 100 languages, it wasn't optimized for multi-lingual models — most of the vocabulary isn't shared between languages and therefore the shared knowledge is limited. To overcome that, XLM (Lample and Conneau, 2019) modifies BERT in two ways including uses Byte-Pair Encoding (BPE) that splits the input into the most common sub-words across all languages, thereby increasing the shared vocabulary between languages. Also, it upgrades the BERT architecture in two manners namely (i) including training sample consists of the same text in two languages, whereas in BERT each sample is built from a single language and (ii) use of context from one language to predict tokens in the other language.

5. **ALBERT:** ALBERT (Lan et al., 2019) is an extension of BERT with more focus on reduction in architecture with Factorized embedding parameterization, Cross-layer parameter sharing and Inter-sentence coherence loss.

6. **DistillBERT:** DistillBERT (Sanh et al., 2019) again similar to ALBERT focuses on model size reduction, except uses knowledge distillation.

7. **Term Frequency (TF):** Term Frequency a.k.a BOW or Term Count measures how frequently a term occurs in a document. Since every document is different in length, multiple tokens would appear much more in long documents than shorter ones. Thus, we divide the term frequency using the document length. The formulae are as shown in

$$\mathsf{TF}(t) = \frac{\text{Number of times term t appears in a document}}{\text{Total number of terms in the document}} \quad (3)$$

8. **Term Frequency Inverse document frequency (TFIDF):** The TFIDF algorithm builds on the following representation of sentences/documents. Each sentence d is represented as a vector $d = (d_1, d_2, ...., d_F)$ so that documents with similar content have similar vectors (according to a fixed similarity metric). Each element $d_i$ represents a distinct word $w_i$. $d_i$ for a sentence d is calculated as a combination of the statistics $TF(w_i; d)$ and $DF(w_i)$. The term frequency $TF(w_i; d)$ is the number of times word $w_i$ occurs in document d and the document frequency $DF(w_i)$ is the number of documents in which word $w_i$ occurs at least once. The inverse document frequency IDF $(w_i)$ can be calculated from the document frequency.

$$\mathsf{IDF}(w_i) = \log\left(\frac{D}{DF(w_{(i)})}\right) \quad (4)$$

Here, D is the total number of sentences. Intuitively, the inverse document frequency of a word is low if it occurs in many documents and is highest if the word occurs in only one. The so-called weight d(i) of word $w_i$ in sentence $d^i$ is then given as

$$d^{(i)} = TF(w_i; d) IDF(w_i) \tag{5}$$

This word weighting heuristic says that a word $w_i$ is an important indexing term for document d if it occurs frequently in it (the term frequency is high). On the other hand, words that occur in many documents are rated less important indexing terms due to their low inverse document frequency. In this work, we use TF-IDF with Bi-Grams of words.

## 5.2 Classification Algorithm

1. **Support Vector Machines (SVM):** Support Vector Machine (Steinwart and Christmann, 2008) is an classification algorithm used to find a hyperplane in an N-dimensional space that distinctly classifies the data points by maximizing the margin between the data points and the hyperplane.

2. **EXtreme Gradient Boosted Machines (XGBoost):** XGBoost (Chen and Guestrin, 2016) is a decision-tree-based ensemble Machine Learning algorithm that uses a gradient boosting framework where it approaches the process of sequential tree building using parallel implementation and uses cache awareness for hard ware optimization by allocating internal buffers in each thread to store gradient statistics.

## 5.3 Topic Modeling

For topic modeling, we use only Latent Dirichlet Allocation.

1. **Latent Dirichlet Allocation (LDA):** Originally introduced by works of (Blei, Ng, and Jordan, 2003) is a model that allows explaining of observation through the usage of unobserved groups. For example, if observations are words collected into documents, it posits that each document is a mixture of a small number of topics and that each word's presence is attributable to one of the document's topics.

## 5.4 Clustering Algorithms

1. **K-Means:** The k-means algorithm is used to partition a given set of observations into a predefined amount of k clusters. The algorithm starts with a random set of center data points and then updates by assigning observations are assigned to the nearest center. If multiple centers have the same distance to the observation, a random one would be chosen. Following this, the centers are repositioned by calculating the mean of assigned observations to respective center points. The update process reoccurs until all observations remain at the assigned center-points and therefore the center-points would not be updated anymore. K-Means is sensitive to initial centroids where it could end up splitting common data points while other data points get grouped. Further, some of the points are more attracted to outliers. Hence we test three different types of initialization namely K-Means++, Random and PCA components with maximal variance

2. **Aggolomerative Hierarchical Clustering:** This algorithm works by grouping the data one by one based on the nearest distance measure of all the pairwise distance between the data point. Again distance between the data point is recalculated but which distance to consider when the groups have been formed. For this, there are many available methods. Some of them are *single linkage, complete linkage, average linkage, centroid distance, and ward's distance.* In this work, we use the ward's method and group the data until one cluster is formed. We further use three methods for affinity (distance computation for linkage approaches) namely Euclidean, Cosine & City Block Distance.

3. **Spectral Clustering:** Spectral clustering, works by treating the data points as nodes of a graph and clustering is treated as graph partitioning problem. Finally, the nodes are mapped to low-dimensional space to form the final set of clusters. For assigning the low dimension points, the algorithm uses either K-Means or Discretization. In this work, we test both the approaches with spectral clustering.

## 5.5 Supervised Keyphrase Extraction Algorithms

1. **KEA:** KEA (Witten et al., 1999) keyphrase extraction algorithm is a supervised approach for retrieval of important phrases from the text. Originally, KEA uses the Naive Bayes Machine Learning Algorithm for training a binary classifier where the phrases that are used with a dataset that is cleaned linguistically by removing stop words, proper names, case-folding, stemming, etc. Each of these sentences is then converted into a TF-IDF matrix as shown earlier. KEA doesn't use any controlled vocabulary instead uses keyphrases from the input text itself. Additionally, the usage of TF-IDF shows that KEA indeed uses lexical information to extract and characterize the phrases from the document.

2. **WINGNUS:** WINGNUS (Nguyen and Luong, 2010), is similar to KEA except it uses document structure to mine the required keyphrases. WINGUS similar to KEA uses Naive Bayes classifier with TF-IDF, word-offset, phrase length, typeface attribute, title information, title overlap, features indicating phrase appearance and appeerence frequencies in Header, Abstract, Introduction, and other sections of documents.

### 5.5.1 *Unsupervised Keyphrase Extraction Algorithms*

1. **KPMINER:** KPMINER (El-Beltagy and Rafea, 2010) uses a three-step process involving keyphrase selection, weight calculation, and refinement respectively. Candidate keyphrase selection is done using series of rules predominantly seen in textual documents including separation by punctuation mark without any stop words within a given candidate, this is followed by minimal phrase appearance characteristic represented by least allowable seen frequency factor (n) and finally, a so a cutoff constant (*CutOff*) is defined in terms of several words after which if a phrase appears for the first time, it is filtered out and ignored. This will return a series of candidates which are weighed using TF-IDF supplemented with boosting factors. Finally, the weighed candidate phrases are refined with an input keyphrase selection parameters. Overall KPMINER uses document-related information for elucidating the keyphrases.

2. **YAKE:** YAKE (Campos et al., 2020) is a light-weight unsupervised automatic keyword extraction method which rests on statistical text features extracted from single documents to select the most relevant keywords of a text. YAKE is domain-independent, data-independent and free of TF-IDF usually seen in supervised methods. YAKE consists of five major steps namely (1) text preprocessing and candidate term identification - similar to supervised keyphrase extraction, it preprocesses documents (2) feature extraction - uses a set of statistical features to represent the candidates. Typically used features include TF, TF - upper case letters, co-occurrence matrix, sentence offsets, etc. (3) computing term score - these features are heuristically combined into a single score likely to reflect the importance of the term (4) n-gram generation and computing candidate keyword score - generates n-grams and assigns importance scores and (5) data de-duplication and ranking - Finally, terms are ranked based on importance through de-duplication distance similarity measure.

## 5.6 Graph based Ranking Algorithms

1. **TextRank:** TextRank (Mihalcea and Tarau, 2004) constructs graph representations for the input text data following which a graph-based ranking algorithm is then applied to extract the important lexical units (words) in the text. TextRank as part of its implementation uses words as nodes with lexical information such as art-of-speech (nouns and adjectives) and edges are co-occurrence relations, which are managed through word occurrence distances. Following this, the nodes are ranked using an unweighted graph-based ranking algorithm where the iteration of the graph-based ranking algorithm is done until convergence and vertices are sorted based on final scores. Finally, the values attached to each vertex for ranking/selection decisions.

2. **SingleRank:** SingleRank (Wan and Xiao, 2008) originally employs the clustering algorithm to group the documents into a few clusters. The documents within each cluster are expected to be topic-related and each cluster can be considered as a context for any document in the cluster. For each of the cluster, it executes two major steps, first, it does cluster level word evaluation where it builds an affinity graph similar to TextRank followed by cluster-level saliency score is calculated for each word using graph ranking. Following this candidate phrases in the document based on the scores of the words contained in the phrases are evaluated to choose final keyphrases.

3. **TopicRank:** TopicRank (Bougouin, Boudin, and Daille, 2013) is similar to SingleRank where it uses clustering for selecting representative candidates where the document is pre-processed (sentence segmentation, word tokenization, and Part-of-Speech tagging) and keyphrase candidates are clustered into topics. Then, topics are ranked according to their importance in the document and keyphrases are extracted by selecting one keyphrase candidate for each of the most important topics.

4. **PositionRank:** PositionRank (Florescu and Caragea, 2017) algorithm involves three essential steps: (1) the graph construction at word level similar to TextRank (2) the design of Position-Biased PageRank - where unlike page rank PositionRank is to assign larger weights (or probabilities) to words that are found early in a document and are frequent. (3) the formation of candidate phrases where Candidate words that have contiguous positions in a document are concatenated into phrases.

5. **MultipartiteRank:** MultipartiteRank (Boudin, 2018) extends PositionRank, by

seleting keyphrase candidates from the sequences of adjacent nouns with one or more preceding adjectives. They are then grouped into topics based on the stem forms of the words they share using hierarchical agglomerative clustering with average linkage.

## 6 RESULTS, EXPERIMENTS & DISCUSSIONS

In this section, we present results for each of the experimental results and analysis for research questions from section 1.

### 6.1 RQ 1: Keyphrase Extraction

#### 6.1.1 RQ 1.1: Supervised Keyphrase Extraction

Our first research question concentrates on the effectiveness of supervised keyphrase extraction approaches in complex assignments. In this work, we test two approaches, namely KEA and WINGNUS for supervised keyphrase extraction (Section 5.5) by examining Precision (P), Recall (R) and F1 metrics under the setting of both weighed average and macro average (See Table 4). The weighted average is to make results analogous across methods due to an imbalance in data. Both supervised approaches of KEA and WINGNUS internally use TF-IDF as a feature extractor (See sections 5.1) with the Naive Bayes classifier. Additionally, the latter approach uses document related features (See section 5.5).

First look at the results in Table 4 we can see the WINGNUS surmounts KEA by 2% and 4% in F1. With WINGNUS, the performance is comparatively higher because of the use of document and phrase-level features.

*Table 4*—Results of phrase classification using supervised, unsupervised and graph-based approaches.

|  | Fold 1 | | | | | | | Fold 2 | | | | | | |
|---|---|---|---|---|---|---|---|---|---|---|---|---|---|---|
|  |  | Macro Avg | | | Weighed Average | | |  | Macro Avg | | | Weighed Average | | |
|  | Accuracy | P | R | F | P | R | F | Accuracy | P | R | F | P | R | F |
| KEA | 0.58 | 0.61 | 0.62 | 0.58 | 0.68 | 0.58 | 0.59 | 0.50 | 0.5 | 0.25 | 0.33 | 1.00 | 0.50 | 0.67 |
| WINGNUS | 0.61 | 0.61 | 0.62 | 0.60 | 0.67 | 0.61 | 0.63 | 0.58 | 0.5 | 0.29 | 0.37 | 1.00 | 0.58 | 0.73 |
| KP MINER | 0.51 | 0.54 | 0.55 | 0.51 | 0.60 | 0.51 | 0.52 | 0.83 | 0.5 | 0.41 | 0.45 | 1.00 | 0.83 | 0.90 |
| YAKE | 0.62 | 0.57 | 0.58 | 0.57 | 0.62 | 0.62 | 0.62 | 0.94 | 0.5 | 0.47 | 0.48 | 1.00 | 0.94 | 0.97 |
| TOPIC RANK | 0.64 | 0.58 | 0.58 | 0.58 | 0.63 | 0.64 | 0.63 | 0.75 | 0.5 | 0.38 | 0.43 | 1.00 | 0.75 | 0.86 |
| TEXT RANK | 0.35 | 0.61 | 0.51 | 0.29 | 0.70 | 0.35 | 0.22 | 0.85 | 0.5 | 0.42 | 0.46 | 1.00 | 0.85 | 0.92 |
| SINGLE RANK | 0.35 | 0.58 | 0.51 | 0.28 | 0.67 | 0.35 | 0.20 | 0.57 | 0.5 | 0.28 | 0.36 | 1.00 | 0.57 | 0.72 |
| POSITION RANK | 0.45 | 0.56 | 0.55 | 0.45 | 0.62 | 0.45 | 0.43 | 0.90 | 0.5 | 0.45 | 0.47 | 1.00 | 0.90 | 0.95 |
| MULTIPARTITE RANK | 0.67 | 0.63 | 0.64 | 0.64 | 0.68 | 0.67 | 0.67 | 0.94 | 0.5 | 0.48 | 0.48 | 1.00 | 0.94 | 0.97 |

Analyzing results across folds, we can see that fold-1 results are lower than fold-2. We can attribute this to the fold-2 creation process where we ensured that fold-2 has a higher overlap of vocabulary compared to fold-1. (Section 4). However, more thorough linguistic analysis is warranted to understand the impact of specific words on the classification of keyphrases and non-keyphrases. We leave such a study for our future work. In fold-1 we can see even precision and recall, meanwhile, in fold-2 the precision is always 0.5, this is because in fold-2 both the algorithms cannot identify the non-keyphrases. To understand this in more details consider Table 5.

*Table* 5—Results of KEA algorithm with class-wise separation.

|         |                 | Fold 1 |      |      | Fold 2 |      |      |
|---------|-----------------|--------|------|------|--------|------|------|
|         |                 | P      | R    | F1   | P      | R    | F1   |
| Classes | Non-keyphrases  | 0.42   | 0.76 | 0.54 | 0      | 0    | 0    |
|         | Keyphrases      | 0.8    | 0.49 | 0.58 | 1      | 0.5  | 0.66 |
| Metrics | Macro Avg       | 0.61   | 0.62 | 0.58 | 0.5    | 0.25 | 0.33 |
|         | Weighed Average | 0.68   | 0.58 | 0.59 | 1      | 0.5  | 0.67 |

Table 5 shows the results of the KEA algorithm on both fold-1 and fold-2. As visible in Table 5, the fold-2 results drop severely for the classification of non-keyphrases as mentioned earlier, we observe similar behavior for all the algorithms (both supervised and unsupervised). As such, fold-2 warrants a more detailed study with both the algorithms to understand the reason for such behavior. However, we plan to visit this in our future work.

To summarise, our findings are:

- F1 is highest for WINGNUS compared to KEA, with former generating F1 0.60 and 0.37 across Folds 1 and 2, respectively. WINGNUS offers more select performance owing to its usage of the document and phrase-level features besides TF-IDF.
- Across the folds, the performance is weaker in fold-2, with everything identified as a keyphrase. We need more investigation to analyze the root cause. We face this issue again in section 6.1.2.
- Overall, both KEA and WINGNUS serve as encouraging baselines for keyphrase extraction from complex assignments.

### 6.1.2 *RQ 1.2: Unsupervised Keyphrase Extraction*

Unsupervised approaches for keyphrase extraction typically include graph-based ranking and frequency factoring. We comprehensively study an exhaustive set of methods covering both these categories. Table 5 shows the results of KPMINER, YAKE, and series of graph-based ranking approaches. The results vary across the methods ranging from 0.29 SingleRank to 0.64 on multipartite ranking. F1 is lower in SingleRank and TextRank approaches, meanwhile across the rest of the approaches we see the result to be well balanced. Overall, we can see that the results are higher than the supervised approaches.

To begin with, let's consider YAKE and KPMINER approaches. From Table 4 we can see that YAKE delivers better results than KPMINER with 6% and 3% higher F1 than KPMINER approaches. This may be because KPMINER employs simple features such as word frequency factors while YAKE utilizes affinity graph and graph-based ranking. Further in line with results from section 6.1, we can see P of 0.5 and comparatively lower recall.

Meanwhile, the results of the graph-based ranking method cover the entire spectrum, with multipartite rank producing the best results for both fold-1 and fold-2. Again analyzed over fold-2, fold-1 produces better results. Also across the graph-based approaches, we can see that SingleRank delivers the worst, despite both SingleRank and MultipartiteRank using clustering and graph-based modeling approach. We believe this is because of the candidate selection strategy where unlike SingleRank multipartite rank considers candidates with sequences of nouns and adjectives. We can induce similar intuition for the results of SingleRank in fold-1. Overall, we can see that there is no clear favorite in ranking based approaches across both folds 1 and 2. Finally, we can also see that multipartite produces the best accuracy score across all the methods including supervised algorithms.

Coming to the results of fold-2 using graph-based approaches, we can see fixed P values and lower R and F1 values. An in-depth inquiry highlights the same problem encountered earlier in case of supervised approaches, where again the unsupervised approaches perform worse for non-keyphrases (See Table 6).

Finally, all of our results were presented by selecting optimal top K candidate keyphrases identified by the graph-based ranking algorithm. The value of K has a significant influence on overall performance. Figures 3 and 4 shows the impact of various values of K on overall F1 from which we can see that the result is different for fold-1 and fold-2. In the case of fold-1, we can see that algorithms such as PositionRank and MultipartiteRank show high results with lower K and vice versa. Meanwhile, the rest of the algorithms show greater performance with a higher value of K. In contradiction, we can see that in fold-2 all the algorithms saturate around K of 200. Still, we can see that comparable to fold-1 PositionRank and MultipartiteRank produce a high result in fold-2.

*Table 6*—Results of Multipartite algorithm with class-wise separation.

|  |  | Fold 1 | | | Fold 2 | | |
|---|---|---|---|---|---|---|---|
|  |  | P | R | F | P | R | F |
| Classes | Non-keyphrases | 0.50 | 0.54 | 0.52 | 0 | 0 | 0 |
|  | Keyphrases | 0.70 | 0.95 | 0.81 | 1 | 0.94 | 0.96 |
| Metrics | Macro Avg | 0.68 | 0.64 | 0.64 | 0.5 | 0.47 | 0.48 |
|  | Weighed Average | 0.68 | 0.67 | 0.67 | 1 | 0.94 | 0.96 |

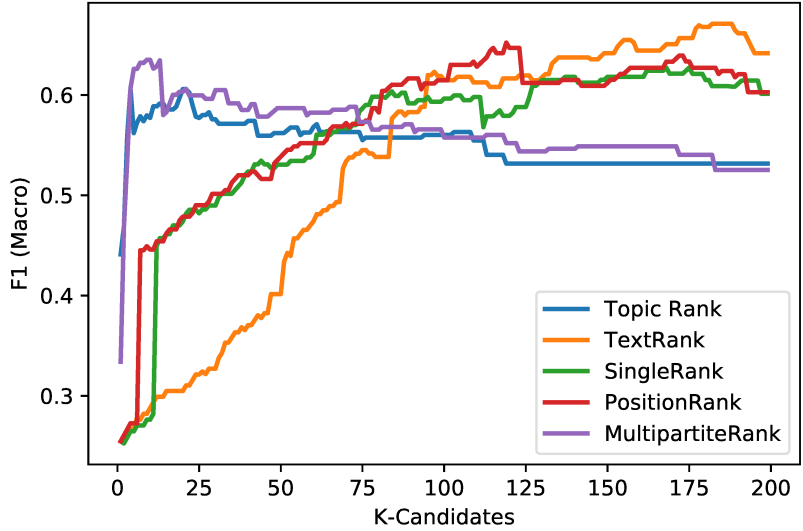

*Figure 3*—Analysis of the impact of number of candidates phrases K on Fold-1.

To summarize we find

- MultipartiteRank provides best results across all the methods both supervised/other unsupervised methods with F1 of 0.64 and 0.48 across folds 1 and 2.
- KPMINER and YAKE offer comparable performances, with YAKE producing

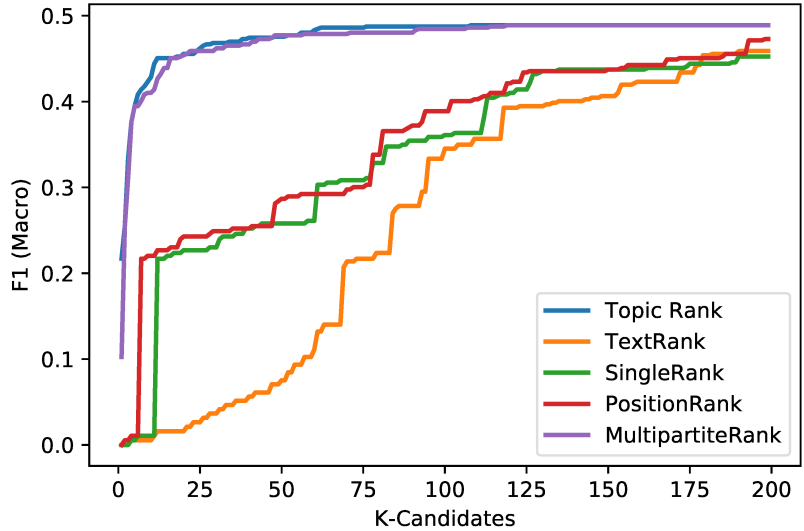

*Figure 4*—Analysis of the impact of number of candidates phrases K on Fold-2.

superior F1 of 0.57 and 0.48 across the two folds.

- Similar to supervised keyphrase extraction, we see that in fold-2 unsupervised approaches again cannot identify any of the non-keyphrases, which in turn warrants deeper analysis of the fold-2 dataset. We plan to revisit this in the future.
- Finally, from Figures 3 and 4 we can see the choice of best candidates influences results and relative impact is conditional on the folds of the dataset.

### 6.2 RQ 2: Specific Keyphrase-Rubric Relationship Extraction

In this section, we will concentrate on the analysis of unsupervised and supervised methods for specific keyphrase-rubric relationship extraction.

#### 6.2.1 RQ 2.1: Unsupervised Approaches

Originally, clustering uses labeled phrases to capture the context of phrases and adapts unlabeled phrases to its centroids. Thus the category of each phrase cluster is identified by the majority labels of texts in it. Hence here clustering is used to generate the classification.

Our first research question in specific keyphrase-rubric relationship classification focuses on the analysis of the effectiveness of clustering approaches for phrases-rubric relationship classification. To elucidate this, Precision (P), Recall

(R) and F1 metrics are reported on both as weighed average level and original macro-level (See Tables 7-9). Moreover, we present additional metrics such as *cluster purity (CP), Rand Index (RI), Cluster Silhouette (SIL)* to understand the nature of clusters. The results obtained for both the folds across multiple unsupervised methods are as shown in Tables 7-9.

*Table 7*—Consolidated results of K-means clustering

| Split | Initialization | Feature Extractor | Accuracy | Macro Average | | | Weighed Average | | | Clustering Metrics | | | |
|---|---|---|---|---|---|---|---|---|---|---|---|---|---|
| | | | | P | R | F1 | P | R | F1 | CP | RI | AMI | SIL |
| Fold 1 | K-Means++ | TF-IDF | 0.3697 | 0.3502 | 0.34701 | 0.31755 | 0.47341 | 0.36957 | 0.39493 | 0.019 | -0.005 | 0.005 | 0.002 |
| | | BERT | 0.38043 | 0.32827 | 0.33661 | 0.28713 | 0.5001 | 0.38043 | 0.40216 | 0.027 | 0.028 | 0.013 | 0.226 |
| | Random | TF-IDF | 0.27174 | 0.21436 | 0.19521 | 0.19884 | 0.31124 | 0.27174 | 0.28683 | 0.027 | -0.006 | 0.012 | 0.002 |
| | | BERT | 0.17391 | 0.37754 | 0.30881 | 0.16963 | 0.59236 | 0.17391 | 0.1274 | 0.023 | 0.011 | 0.009 | 0.205 |
| | PCA | TF-IDF | 0.20652 | 0.20872 | 0.20747 | 0.17234 | 0.3022 | 0.20652 | 0.20179 | 0.019 | 0.011 | 0.005 | 0.003 |
| | | BERT | 0.11413 | 0.11156 | 0.20486 | 0.09994 | 0.15941 | 0.11413 | 0.1024 | 0.024 | 0.026 | 0.01 | 0.234 |
| Fold 2 | K-Means++ | TF-IDF | 0.36413 | 0.32046 | 0.33986 | 0.31261 | 0.43442 | 0.36413 | 0.3811 | 0.042 | 0.027 | 0.026 | 0.003 |
| | | BERT | 0.35326 | 0.32407 | 0.29857 | 0.27175 | 0.39 | 0.35326 | 0.34836 | 0.023 | 0.03 | 0.008 | 0.247 |
| | Random | TF-IDF | 0.2663 | 0.28733 | 0.26625 | 0.24597 | 0.37864 | 0.2663 | 0.29987 | 0.035 | 0.012 | 0.02 | 0.003 |
| | | BERT | 0.33696 | 0.2485 | 0.25248 | 0.23594 | 0.38003 | 0.33696 | 0.34856 | 0.034 | 0.035 | 0.02 | 0.232 |
| | PCA | TF-IDF | 0.2663 | 0.19071 | 0.1972 | 0.1871 | 0.28021 | 0.2663 | 0.26773 | 0.044 | 0.009 | 0.029 | 0.004 |
| | | BERT | 0.20652 | 0.1636 | 0.20035 | 0.15224 | 0.254147 | 0.20652 | 0.21426 | 0.023 | 0.007 | 0.008 | 0.969 |

From Table 7, collating all the results we can see that K-Means produces the best result, followed by Spectral clustering and then Agglomerative. More specifically, while K-Means produces the best result of 0.32 F1-Score Agglomerative clustering produces high precision and recall of 0.48 the macro F1 is lower than 0.2. Further, the behavior is consistent across both the folds. Also, we see high precision in Fold-2 and a balanced P and R in Fold-1. Besides in fold-1, we see high accuracy with Spectral clustering with an accuracy score of 0.45 with F1 of Agglomerative & Spectral methods very far from those of K-Means. Concerning feature extractors, we can see that while TF-IDF produces the top results across the majority of the experiments.

About K-Means, we tested with three different initialization namely *K-Means++, Random & PCA* based initialization. We hypothesize that with multiple random tests, the random initialization would provide results similar to K-Means++ and principal components with maximum variance indeed produces useful results. Results so obtained across both the folds are again presented in Table 7. We can see that K-means++ offers results higher than the Random and PCA based initialization. Additionally, we can also see that random initialization, yields results very close to that of K-Means++. Meanwhile, the maximal variance components are indeed weak and biased on the identification of the phrase level classes.

*Table 8*—Consolidated results of Agglomerative Clustering

| Split | Affinity | Feature Extractor | Accuracy | Macro Average | | | Weighed Average | | | Clustering Metrics | | | |
|---|---|---|---|---|---|---|---|---|---|---|---|---|---|
| | | | | P | R | F1 | P | R | F1 | CP | RI | AMI | SIL |
| Fold 1 | Cosine | TF-IDF | 0.26087 | 0.22599 | 0.22488 | 0.19884 | 0.2601 | 0.26087 | 0.2254 | 0.041 | 0.093 | 0.02 | 0 |
| | | BERT | 0.25 | 0.18224 | 0.21951 | 0.15287 | 0.25389 | 0.25 | 0.1738 | 0.019 | 0.008 | -0.006 | 0.002 |
| | Euclidean | TF-IDF | 0.2663 | 0.19418 | 0.20472 | 0.18799 | 0.28933 | 0.2663 | 0.26072 | 0.034 | 0.013 | 0.011 | 0.002 |
| | | BERT | 0.28261 | 0.20072 | 0.19133 | 0.18597 | 0.30854 | 0.28261 | 0.28316 | 0.035 | 0.015 | 0.013 | 0.004 |
| | Cityblock | TF-IDF | 0.33152 | 0.49767 | 0.26597 | 0.15312 | 0.55866 | 0.33152 | 0.18576 | 0.021 | 0.03 | 0.004 | -0.023 |
| | | BERT | 0.30435 | 0.57714 | 0.25472 | 0.14321 | 0.70804 | 0.30435 | 0.16346 | 0.026 | 0.021 | 0.011 | -0.023 |
| Fold 2 | Cosine | TF-IDF | 0.26087 | 0.1803 | 0.24574 | 0.16725 | 0.26031 | 0.26087 | 0.1971 | 0.014 | 0.034 | -0.008 | 0.222 |
| | | BERT | 0.29348 | 0.29348 | 0.24579 | 0.13554 | 0.26948 | 0.29348 | 0.15777 | 0.016 | 0.003 | -0.008 | 0.204 |
| | Euclidean | TF-IDF | 0.33152 | 0.20736 | 0.23688 | 0.18022 | 0.35104 | 0.33152 | 0.2753 | 0.019 | 0.013 | -0.002 | 0.169 |
| | | BERT | 0.23913 | 0.48054 | 0.48054 | 0.18138 | 0.4694 | 0.23913 | 0.16654 | 0.022 | 0.011 | 0.002 | 0.223 |
| | Cityblock | TF-IDF | 0.42935 | 0.26665 | 0.28221 | 0.27278 | 0.41755 | 0.42935 | 0.42126 | 0.033 | 0.032 | 0.014 | 0.182 |
| | | BERT | 0.21196 | 0.26006 | 0.2676 | 0.20384 | 0.3732 | 0.21196 | 0.23477 | 0.031 | -0.008 | 0.009 | 0.353 |

In the case of Agglomerative Clustering, we adopted ward distance for linkage and multiple affinity measure namely cosine, euclidean and city block distance metrics. Table 8 displays the consolidated results so obtained. From Table 8 we can see that Euclidean Distance gives the best results (similar to original wards computation). Cosine distance yields similar results like that of Euclidean under TF-IDF feature extractor but in the rest of the cases, Euclidean outperforms City block distance which offers the worst performance. Overall agglomerative clustering produces best results of 0.18 F1 using Euclidean affinity on BERT features.

*Table 9*—Consolidated Results of Spectral Clustering

| Split | Sampling Strategy | Feature Extractor | Accuracy | Macro Average | | | Weighed Average | | | Clustering Metrics | | | |
|---|---|---|---|---|---|---|---|---|---|---|---|---|---|
| | | | | P | R | F1 | P | R | F1 | CP | RI | AMI | SIL |
| Fold 1 | K-Means | TF-IDF | 0.41304 | 0.35258 | 0.27234 | 0.2713 | 0.41887 | 0.41304 | 0.39264 | 0.036 | 0.002 | 0.016 | 0.01 |
| | | BERT | 0.45109 | 0.34087 | 0.30922 | 0.28039 | 0.42579 | 0.45109 | 0.40373 | 0.038 | -0.034 | 0.02 | 0.01 |
| | Discrete | TF-IDF | 0.21196 | 0.36059 | 0.2215 | 0.16441 | 0.60518 | 0.21196 | 0.18973 | 0.04 | 0 | 0.019 | 0.008 |
| | | BERT | 0.2663 | 0.181 | 0.23257 | 0.15539 | 0.25327 | 0.2663 | 0.17572 | 0.016 | -0.023 | -0.007 | 0.009 |
| Fold 2 | K-Means | TF-IDF | 0.28804 | 0.29388 | 0.23026 | 0.16994 | 0.50693 | 0.28804 | 0.23666 | 0.036 | 0.002 | 0.016 | 0.01 |
| | | BERT | 0.29348 | 0.075 | 0.24107 | 0.11441 | 0.0913 | 0.29348 | 0.13928 | 0.024 | 0.007 | 0.012 | 0.393 |
| | Discrete | TF-IDF | 0.28261 | 0.31993 | 0.31344 | 0.24516 | 0.46287 | 0.28261 | 0.28912 | 0.04 | 0 | 0.019 | 0.008 |
| | | BERT | 0.45652 | 0.20798 | 0.24683 | 0.20448 | 0.34264 | 0.45652 | 0.36264 | 0.026 | -0.02 | 0.009 | 0.257 |

Finally, in spectral clustering (See Table 9) the behavior is similar to that of K-Means, because internally spectral clustering, in turn, uses K-means. The best performing spectral clustering approach gives an F1-Score of 0.28 with BERT features similar results could be seen even with TF-IDF feature extraction. K-Means sample selection produces an accuracy score of 0.451 F1 while discretized hierarchical clustering gives the best result of 0.16 F1.

The results of cluster analysis are as shown in the cluster metrics including cluster purity, rand index, etc. We can see that across all the metrics the clusters are not so good. Much of the cluster doesn't satisfy homogeneity indicating that clusters don't contain only data points which are members of a single class. This is also visible through low P, R and F1 measures. While we can further see the average mutual information is very low indicating the clusters are completely mixed. Additionally, we can see that the silhouette is very low for all the experiments this shows that clusters are extremely overlapping. To summarise, our findings are:

- F1 is highest for K-Means with K-Means++ initialization. The findings also hold for spectral clustering.
- Tf-IDF representation produces the highest results majority of the approaches compared to BERT.
- Agglomerative clustering produces the least results. The same is true with spectral clustering.
- Compared to both the Folds, fold-1 shows better performance than fold-2.
- From cluster analysis, we can see that clusters are overlapping with near-zero silhouette score and homogeneity.

### 6.2.2 RQ 2.2: Supervised & Topic Modeling Approaches

Previously in section 6.2.1, we reviewed unsupervised models on phrase-rubric relationship extraction with K-Means ruling the performance with K-Means++ initialization. However textual content usually has multiple topics which in our case is *Task, Reasoning, Intuition, Findings*. Hence in this research question, we will employ topic modeling to see if these methods are indeed useful for the process of relationship classification. Generally, a topic model is a type of statistical model for discovering the abstract "topics" that occur in a collection of documents. In this work, latent Dirichlet's allocation Blei, Ng, and Jordan, (2003) was used. We further contrast performance to supervised classification models involving SVM with TF-IDF and BERT. Precision (P), Recall (R) and F1 metrics are reported again on both as weighed average level and original macro-level. Tables 10 and 11 shows results of LDA and supervised approaches.

In K-means, we used K-Means with TF-IDF features coupled with Latent Dirich-let Allocation. Similar to results from the previous section 6.2.1 we can see that K-Means with K-Means++ and Random initialization performs likewise. Also including LDA improves results for randomly initialized k-means. However, the results are significantly lower than the original implementation with only TF-IDF and K-Means++. Overall LDA contributes to improving accuracy score by 0.06 with a drop in F1 of 0.08 in fold-1 and 0.05 with a drop in F1 by 0.15 in fold-2. Overall, we can see an improvement in accuracy score across all the initialization post introduction of LDA. To recognize which among the rubrics are easily identifiable we present confusion matrix in 5. We can see that cognitive connection is easy to map among all the rubrics, this is because of a limited vocabulary of the sentences used to identify the cognitive connection, while the rest of the rubrics are difficult to determine.

*Table 10*—Consolidates results using Latent Dirichlet Allocation with Clustering Algorithms

|  |  |  | | Macro Average | | | Weighed Average | | | Clustering Metrics | | | |
|---|---|---|---|---|---|---|---|---|---|---|---|---|---|
| Split | Approach | Parameter | Accuracy | P | R | F | P | R | F | CP | RI | AMI | SIL |
| Fold 1 | K-Means | K-Means++ | 0.43 | 0.27 | 0.26 | 0.24 | 0.38 | 0.43 | 0.36 | 0.03 | 0.04 | 0.01 | 0.98 |
| | | Random | 0.43 | 0.27 | 0.26 | 0.24 | 0.38 | 0.43 | 0.36 | 0.03 | 0.04 | 0.01 | 0.98 |
| | | PCA | 0.14 | 0.17 | 0.20 | 0.10 | 0.28 | 0.14 | 0.11 | 0.03 | 0.04 | 0.01 | 0.98 |
| | Agglomerative | Cosine | 0.30 | 0.20 | 0.25 | 0.17 | 0.27 | 0.30 | 0.19 | 0.04 | -0.01 | 0.02 | 0.48 |
| | | Euclidean | 0.33 | 0.33 | 0.26 | 0.18 | 0.51 | 0.33 | 0.25 | 0.04 | -0.01 | 0.02 | 0.48 |
| | | Cityblock | 0.32 | 0.31 | 0.25 | 0.17 | 0.45 | 0.32 | 0.23 | 0.03 | -0.01 | 0.00 | 0.48 |
| | Spectral | K-Means | 0.26 | 0.26 | 0.22 | 0.17 | 0.35 | 0.26 | 0.22 | 0.02 | 0.01 | -0.01 | 0.47 |
| | | Discrete | 0.35 | 0.27 | 0.28 | 0.26 | 0.38 | 0.35 | 0.35 | 0.01 | 0.01 | -0.01 | 0.46 |
| Fold 2 | K-Means | K-Means++ | 0.21 | 0.16 | 0.20 | 0.15 | 0.25 | 0.21 | 0.21 | 0.02 | 0.01 | 0.01 | 0.97 |
| | | Random | 0.35 | 0.29 | 0.32 | 0.29 | 0.38 | 0.35 | 0.36 | 0.02 | 0.01 | 0.01 | 0.97 |
| | | PCA | 0.27 | 0.28 | 0.29 | 0.25 | 0.38 | 0.27 | 0.27 | 0.02 | 0.01 | 0.01 | 0.97 |
| | Agglomerative | Cosine | 0.20 | 0.24 | 0.17 | 0.16 | 0.36 | 0.20 | 0.21 | 0.04 | 0.01 | 0.02 | 0.39 |
| | | Euclidean | 0.28 | 0.26 | 0.29 | 0.24 | 0.34 | 0.28 | 0.29 | 0.05 | 0.02 | 0.02 | 0.35 |
| | | Cityblock | 0.21 | 0.26 | 0.27 | 0.20 | 0.37 | 0.21 | 0.23 | 0.03 | -0.01 | 0.01 | 0.35 |
| | Spectral | K-Means | 0.37 | 0.30 | 0.33 | 0.30 | 0.40 | 0.37 | 0.38 | 0.05 | 0.01 | 0.03 | 0.41 |
| | | Discrete | 0.27 | 0.25 | 0.29 | 0.25 | 0.31 | 0.27 | 0.28 | 0.05 | 0.01 | 0.03 | 0.42 |

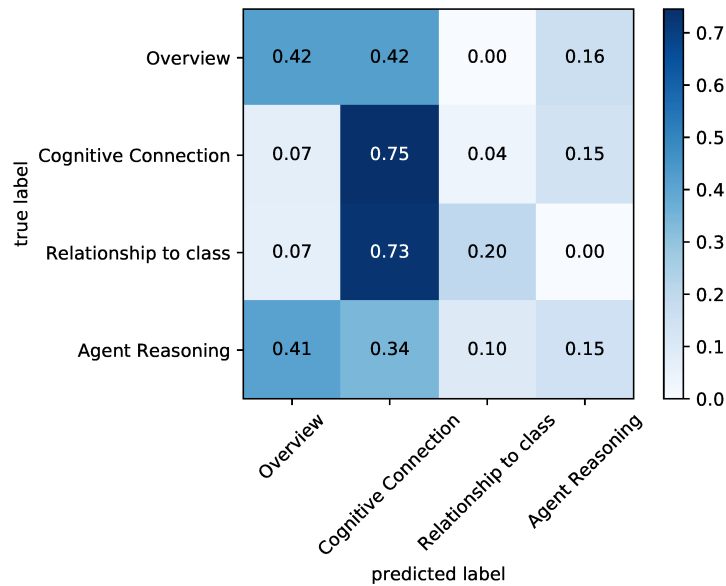

*Figure 5*—Confusion Matrix of the best performing K-Means model

*Table 11*—Results of supervised phrase-rubric relationship classification

| | | Fold 1 | | | | | | Fold 2 | | | | | |
|---|---|---|---|---|---|---|---|---|---|---|---|---|---|
| | | | Macro Average | | | Weighed Average | | | Macro Average | | | Weighed Average | |
| Feature Extractor | Classifier | Accuracy | P | R | F | P | R | F | Accuracy | P | R | F | P | R | F |
| BOW | | 0.54 | 0.46 | 0.41 | 0.42 | 0.53 | 0.54 | 0.52 | 0.54 | 0.38 | 0.37 | 0.37 | 0.51 | 0.54 | 0.52 |
| TFIDF | SVM | 0.57 | 0.44 | 0.40 | 0.41 | 0.53 | 0.56 | 0.54 | 0.59 | 0.45 | 0.41 | 0.42 | 0.64 | 0.59 | 0.61 |
| **BERT** | | **0.56** | **0.47** | **0.40** | **0.41** | **0.57** | **0.56** | **0.52** | **0.60** | **0.51** | **0.50** | **0.48** | **0.63** | **0.60** | **0.61** |
| BERT | XGB | 0.56 | 0.49 | 0.37 | 0.38 | 0.54 | 0.56 | 0.52 | 0.55 | 0.44 | 0.35 | 0.35 | 0.52 | 0.55 | 0.50 |

In the case of agglomerative clustering, the results are very comparable to K-Means, where the accuracy score increased by 0.07 with similar F1-Score as shown in section 6.2.1 for agglomerative clustering. However, we can also see that the city block affinity exhibits higher sensitivity compared to cosine affinity.

In spectral clustering we see that results are again alike to K-Means, however, there is surprising finding where the results for spectral clustering with K-Means++ sampling are lower than that of Discretized Sample selection process. The difference in results is 0.1 in accuracy and 0.08 F1-Score. Overall with LDA discretized sampling shows the best results for Spectral clustering. Similar observations can be seen in fold-2 of the dataset except spectral clustering with K-Means yielding higher results than that of fold-1.

Overall with LDA, we see two major benefits firstly we can see an enhancement in accuracy score. Otherwise overall results are significantly lower. This is because in this work we used total as four topics during LDA calculation however based on results we can assume that the dataset contains topics more than four. The supervised results are as shown in Table 11. We can see the best performing supervised approach outperforms the best-unsupervised approaches by an average of 0.15 F1 across the folds.

Finally, coming to cluster analysis we can see that the cluster overall is signif-icantly decreased as evident by the silhouette score of 0.9 with K-Means, 0.3 with agglomerative and 0.4 with spectral clustering. Yet the cluster uniqueness is fairly inadequate as evident with homogeneity metrics and rand index where the values are close to zero.

To summarise, our findings are:

- Firstly, LDA enhances overall accuracy across all the unsupervised algorithms. Further the results are comparable to that of section 6.2. The maximum F1 in fold-1 is 0.25 and in fold-2 is 0.29 with both using spectral clustering.
- Further, spectral clustering shows a novel behavior where the results flip between discrete and K-Means sample selection. However, the overall performance is inferior compared to section 6.2.
- With LDA we can see the clusters don't overlap as evident by silhouette score. However individual clusters are noisy with data from multiple classes.
- Supervised approaches inherently deliver higher results compared to that of unsupervised approaches.

## 6.3 RQ 3: Generic Keyphrase-Rubric Relationship Classification

In this section we dissect traditional and language models in the context of Generic Keyphrase-Rubric Relationship Classification.

### 6.3.1 *RQ 3.1: Traditional Approaches*

Our first research question for keyphrase classification will concentrate on the investigation of performance through traditional supervised approaches. In this work, we confine ourselves to SVM, due to its usage in benchmarks of short text classification (Zeng et al., 2018). To answer the RQ, Precision (P), Recall (R) and F1 metrics are reported on both as weighed average level and original macro-level (See Table 12). Weighted average in keyphrase classification ensures comparable results by taking data imbalance and its sensitivity on SVM. For weighted average prediction, we weigh the dataset predictions based on the ratio of the sample of a category in the dataset. Also inline with section 1.2 we will analyze both the folds of the dataset. We grid search all the hyperparameters for SVM. Also, the first look on the dataset suggests the classes of a task, finding and intuition typically encompass words in pairs (predominantly verbs), which can help in classification. Hence we also search for optimal n-gram parameters as part of the grid search resulting in n-gram value of (1,2).

Besides, we implement SVM using Gradient Descent Classifier with Hinge loss, rather than native SVM formation to reduce overfitting and randomness in initialization. This further ensures the overall method doesn't overfit with fewer data. To ensure a fair comparison, we also create a baseline classifier that predicts class labels proportional to their distribution in the training set. We again show the results of the baseline classifier in Table 12.

Table 12 displays results with Precision (P), recall (R), and F1- Score (F1). We exhibit results for all the folds with an accuracy score. Comparing the different methods, we realized the highest F1 results among traditional methods with SVM with TFIDF, followed by SVM with BOW and Baseline. SVM with TF-IDF shows high precision in Fold-1 and has a balanced P and R in Fold-2, this is unlike what we saw in keyphrase extraction, where the results were higher for fold-1 and in fold-2 we had issues of classifying non-keyphrases. For SVM with BOW, the results are very close to that of TF-IDF with a small difference of 0.2 in the F1 score. These conclusions are in line with other work reporting the usefulness of SVM in short text classification (Zeng et al., 2018).

Regarding results on each of the folds, SVM with BOW offers similar results in fold-1 but again dominated by SVM with TF-IDF in fold-2 by a large margin (6% F1). However, the overall performance is lower as opposed to that of typical text classification benchmarks, mostly because of the semi-formal nature of the text.

*Table 12*—Consolidated Results for Phrase Classification.

| | | Fold 1 | | | | | | | Fold 2 | | | | | | |
| --- | --- | --- | --- | --- | --- | --- | --- | --- | --- | --- | --- | --- | --- | --- | --- |
| | | | Macro Average | | | Weighed Average | | | | Macro Average | | | Weighed Average | | |
| Feature Extractor | Classifier | Accuracy | P | R | F | P | R | F | Accuracy | P | R | F | P | R | F |
| - | Baseline[3] | 0.4 | 0.29 | 0.3 | 0.29 | 0.39 | 0.4 | 0.38 | 0.34 | 0.25 | 0.24 | 0.24 | 0.36 | 0.34 | 0.35 |
| BOW | | 0.54 | 0.46 | 0.41 | 0.42 | 0.53 | 0.54 | 0.52 | 0.54 | 0.38 | 0.37 | 0.37 | 0.51 | 0.54 | 0.52 |
| TFIDF | | 0.57 | 0.44 | 0.40 | 0.41 | 0.53 | 0.56 | 0.54 | 0.59 | 0.45 | 0.41 | 0.42 | 0.64 | 0.59 | 0.61 |
| BERT | | 0.56 | 0.47 | 0.40 | 0.41 | 0.57 | 0.56 | 0.52 | 0.60 | 0.51 | 0.50 | 0.48 | 0.63 | 0.60 | 0.61 |
| ROBERTA | | 0.28 | 0.16 | 0.29 | 0.16 | 0.27 | 0.28 | 0.26 | 0.11 | 0.05 | 0.23 | 0.07 | 0.02 | 0.11 | 0.03 |
| DISTILLBERT | SVM | 0.60 | 0.46 | 0.39 | 0.39 | 0.58 | 0.50 | 0.56 | 0.53 | 0.41 | 0.39 | 0.37 | 0.55 | 0.53 | 0.51 |
| XLM | | 0.63 | 0.53 | 0.48 | 0.49 | 0.61 | 0.63 | 0.62 | 0.55 | 0.40 | 0.38 | 0.39 | 0.53 | 0.55 | 0.53 |
| XLNET | | 0.55 | 0.45 | 0.42 | 0.43 | 0.55 | 0.55 | 0.55 | 0.57 | 0.52 | 0.40 | 0.41 | 0.56 | 0.57 | 0.54 |
| XLMROBERTA | | 0.31 | 0.07 | 0.25 | 0.11 | 0.09 | 0.31 | 0.15 | 0.48 | 0.12 | 0.25 | 0.16 | 0.23 | 0.48 | 0.32 |
| ALBERT | | 0.48 | 0.40 | 0.40 | 0.37 | 0.52 | 0.48 | 0.47 | 0.59 | 0.55 | 0.43 | 0.44 | 0.58 | 0.59 | 0.56 |
| BERT | | 0.56 | 0.49 | 0.37 | 0.38 | 0.54 | 0.56 | 0.52 | 0.55 | 0.44 | 0.35 | 0.35 | 0.52 | 0.55 | 0.50 |
| ROBERTA | | 0.46 | 0.14 | 0.27 | 0.17 | 0.24 | 0.46 | 0.32 | 0.18 | 0.20 | 0.24 | 0.23 | 0.32 | 0.18 | 0.16 |
| DISTILLBERT | | 0.59 | 0.38 | 0.35 | 0.34 | 0.59 | 0.54 | 0.54 | 0.63 | 0.47 | 0.41 | 0.41 | 0.58 | 0.63 | 0.58 |
| XLM | XGB | 0.58 | 0.42 | 0.34 | 0.33 | 0.54 | 0.58 | 0.51 | 0.61 | 0.40 | 0.41 | 0.40 | 0.56 | 0.61 | 0.58 |
| XLNET | | 0.55 | 0.39 | 0.33 | 0.32 | 0.51 | 0.55 | 0.49 | 0.58 | 0.42 | 0.35 | 0.34 | 0.53 | 0.58 | 0.51 |
| XLMROBERTA | | 0.55 | 0.41 | 0.33 | 0.32 | 0.51 | 0.55 | 0.48 | 0.53 | 0.37 | 0.32 | 0.31 | 0.48 | 0.53 | 0.47 |
| ALBERT | | 0.54 | 0.43 | 0.35 | 0.35 | 0.51 | 0.54 | 0.49 | 0.55 | 0.61 | 0.36 | 0.36 | 0.58 | 0.55 | 0.50 |

Our overall analysis of the traditional method shows that it performs fairly well on the complex assignment phrases, however, we believe owing to sparsity in vocabulary and size of dataset the accuracy is lower than text classification benchmarks. Hence, with larger datasets and different domains, the performance is expected to be constant. To summarise, our findings are:

- F1 is highest with SVM with TF-IDF, followed by SVM with BOW.
- SVM with TF-IDF exceeds baseline classifier[2] by a large margin (>= 16% in F1).
- Among both the folds, fold-2 provides better results then fold-1. This is unlike what we saw in keyphrase extraction (See Table 4)
- Existing hypothesis (Zeng et al., 2018) of SVM with TF-IDF as a competitive baseline for short text classification still holds.

### 6.3.2 RQ 3.2: Language Models

While SVM with TF-IDF offers significant results, the error and ambiguity are still high. This is visible from results in which F1 <0.45. Recently, language models offer a significant benefit where they help to build on existing knowledge gathered from large datasets. Since much of the dataset used in this work is derived from semi-formal reports, we hypothesize such pretrained networks would help in results. Besides, since the dataset is very small, training a language model would lead to significant overfitting. Hence to facilitate this we propose to use the pretrained language models only as feature extractors with SVM and XGBoost (Chen and Guestrin, 2016) classifier. We use XGBoost because of its previous success with language models.

Table 12 presents the results on the multiple language models including *BERT*, *RoBERTa*, *DistillBert*, *Albert*, *XLM*, *XLNET*, *XLMROBERTa* transformer models used as feature extractors. The features are extracted from the second encoder of the transformer[3]. To simplify the analysis we shall consider the only performance of BERT and RoBERTa, due to their extensive usage and generalized results. However, a similar analysis could be drawn for others as well.

Firstly, examining the results of all the language models, we can see that BERT offers a balanced F1 of 0.45 an 0.48 across the folds. We can also see that while in fold-1 XLM produces the highest results, in fold-2 the results are significantly lower. In line with the previous section, we again see that results of fold-2 is lesser than fold-1. Besides this, we can also see that results are lowest for RoBERTa in phase-2. We tested this result of RoBERTa multiple times and found neither programming nor correlation level issues. Additionally, to check the impact of training, we froze all the encoders and trained only the final layers, yet the results were lower than the baseline. Also, as shown in Table 12 results using XGB are similar across all models and unlike SVM, the results of P and R are well balanced.

---

[2] We use scikit-learn DummyClassifier with *stratified* strategy.
[3] This was based on our experimentation with different encoders

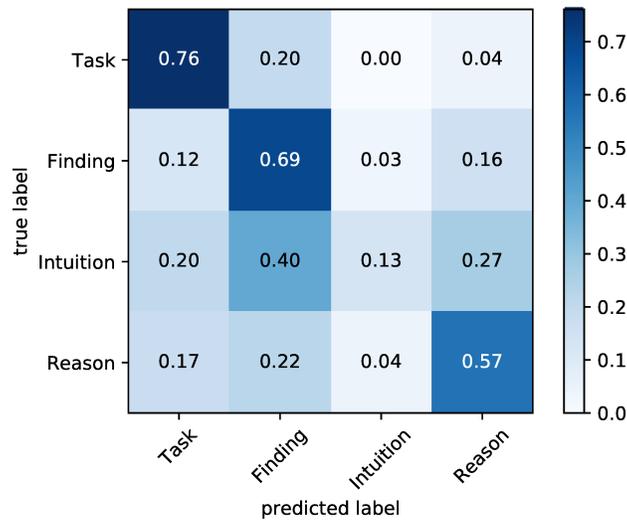

*Figure 6*—Confusion Matrix of the best performing BERT-SVM model

Concerning RoBERTa's performance on SVM in Fold-2 we suspect following, RoBERTa originally uses a special training procedure with dataset larger than BERT, while we expect the pretrained representations to be well distributed, however with training on large sets the representations form tight clusters results in need of greater weighting of data points. However, more experiments are needed for validating the same. The experimental setup of the RoBERTa tokenizer needs a revisit.

Regarding errors, we can see that the classes of *Task, Finding, and Intuition* are highly ambiguous. Figure 6 shows the confusion matrix, where we can see much of the error is concentrated among the three classes. This is in line with Cohen's kappa (κ) of 0.77 earlier mentioned in the dataset.

To summarise, our findings are:

- Results of BERT is higher than that of RoBERTa and best-performing SVM across folds. Meanwhile, XGB offers similar results across different language model representations.
- XLM while delivers the highest results in fold-1, the results are lower with fold-2.
- Despite the sensitivity of an imbalanced dataset, BERT with SVM produces F1 of 0.45 on fold-1, but for RoBERTa we see otherwise. We believe this low result of RoBERTa is because of the inherent architecture. However, such a trait requires more in-depth architectural exploration
- With XGB, both BERT and RoBERTa perform similarly on Fold-1 and Fold-2.
- However, there are still notable differences in the classification of phrases and human annotation. This shows that capturing context is a hard issue, especially with phrases of complex assignments and current annotation schemes. Thus we made need more granular annotation schemes, which we plan to revisit in our future works.

### 6.3.3 *RQ 3.3: Reliability study of traditional and language models*

Previously in Table 12, we saw the results of both traditional and language models. It can be seen that among traditional models the performance is closer, with TF-IDF producing higher performance on fold-2. Meanwhile, BERT language models show significantly higher than the rest. To further test the performance in-depth, we focused on the reliability aspect of the model where we define reliability *as the models' ability to focus on the required information for classification true from a human's perspective* i.e. the model should look on some parts of the text which a human would see during annotation. To do this, we use Local Interpretable Model-agnostic Explanations (LIME) (Ribeiro, Singh, and Guestrin, 2016).

*Table 13*—Results of Interpretability on Traditional and BERT.

|   | Model -1   | Model-2      | Precision |
|---|------------|--------------|-----------|
| 1 | BERT (SVM) | SVM (TF-IDF) | 0.82      |
| 2 | BERT (SVM) | SVM (BOW)    | 0.81      |

We use the **precision** as the evaluation metric, where we measure precision as the fraction of words that are contributing to both traditional and language models performance[4]. The results so obtained are as shown in Table 13. From

---

[4] We calculated precision by manually counting the words.

*Table 14*—Example of Interpretability of Traditional (right) and Language models (left) with similar word contributions.

| TF-IDF (SVM) | | BERT (SVM) | |
|---|---|---|---|
| Feature | Contribution | Feature | Contribution |
| which | +0.350 | agent | +0.13 |
| still can | +0.190 | this | +0.09 |
| problems involving | +0.171 | still | +0.07 |
| have shading | +0.170 | which | +0.04 |
| handle problems | +0.145 | handle | +0.04 |
| involving shapes | +0.137 | problems | +0.03 |
| shapes which | +0.119 | have | +0.01 |
| this agent | +0.079 | can | +0.01 |
| which have | +0.065 | involving | +0.01 |

*Table 15*—Example of Interpretability of Traditional (left) and Language models (right) with dissimilar word contributions.

| TF-IDF (SVM) | | BERT (SVM) | |
|---|---|---|---|
| Feature | Contribution | Feature | Contribution |
| the | +1.154 | a | +0.14 |
| performance | +0.264 | the | +0.13 |
| agent | +0.220 | the | +0.08 |
| performance of | +0.205 | increased | -0.15 |
| a little | +0.074 | has | -0.04 |
| increased | +0.071 | little | -0.03 |
| increased a | +0.055 | of | -0.02 |
| a | -0.051 | performance | -0.02 |
| the agent | -0.089 | agent | -0.01 |

Table 13, we find that in both the cases the models look around 80% of times on similar word lists.

We also manually examine the results to find interesting characteristics. Table 14 shows results where both traditional and language models look on similar words, while Table 15 shows the BERT model to concentrate on words that make little sense in terms of contributions. [5]

From Table 14 it can be seen that both traditional and BERT look for similar words for predicting the correct results, however, from Table 15 we can see that sometimes traditional model has better interpretability. We believe this is an issue of using the BERT model as a feature extractor, rather than training it from scratch.

---

5 We don't use XGB and RoBERTa models as they produce lower performance.

Overall, traditional models focus on the right set of words, in most cases. Hence we conclude that traditional models better interpret the cues related to class from the educational text. However, we argue that this needs to be revisited in more depth with an exhaustive comparison on a relationship with features used in the traditional model and language model training process.

To summarise, our findings are:

- Both traditional and language models view on similar word cues around 80% of times, despite dissimilar contributions.
- Traditional models are more reliable in terms of interpretability, this has more to do with the usage of simple n-gram features unlike language models, complex encoders with no training.

# 7 CONCLUSION

In this work, we focused on keyphrase extraction, generic/specific keyphrase-rubric relationship extraction from complex assignments (section 1). Firstly, in section 4 by developing a new corpus and annotation scheme, we demonstrated that datasets in complex assignments are harder to create: in terms of annotation and size; the balance of classes; the proportion of words; and how often tokens are repeated. The dataset so created is imbalanced with the domination of *Task* and *Finding* classes. This is similar, to the dataset of twitter and web corpora which are traditionally noisy. Also we presented details on how the same annotated dataset is used across all the three tasks in section 4.1.2 along with a brief explanation on various algorithms that was used for the experiments in section 5.

Firstly we began our study on keyphrase extraction in section 6.1 by evaluating supervised approaches of KEA and WINGNUS where WINGNUS offers better performance owing to its usage of the document and phrase-level features besides TF-IDF. Also we saw that the performance is lower in fold-2, with everything identified as a keyphrase. However both KEA and WINGNUS serve as promising baselines for keyphrase extraction from complex assignments. Meanwhile in section 6.1.2 we saw that surprisingly unsupervised approaches offer better results. In unsupervised approaches (section 6.1.2) we tested KPMINER, YAKE and other graph based approaches. We find that MultipartiteRank produces best results across all the methods both supervised/other unsupervised approaches with F1 of 0.64 and 0.48 across folds 1 and 2, meanwhile KPMINER and YAKE produce comparable performances, with YAKE producing superior F1 of 0.57 and 0.48 across the two folds. However, we see that in fold-2 unsupervised approaches again cannot identify any of the non-keyphrases which warrants more thorough algorithmic and linguistic analysis to understand the reason for the drop in performance.

Second, in specific keyphrase-rubric relationship classification (section 6.2) we investigated the capacity of both clustering and topic modeling for the task of classification of phrases-rubrics from complex assignments. We studied three clustering approaches in section 6.2.1 involving K-Means, Agglomerative and Spectral clustering. We find that K-Means to dominate the results with both TF-IDF and BERT against the rest followed by spectral clustering with K-Means based sampling. Agglomerative clustering performs the worst across the three. However the best traditional methods results are significantly lower than that of supervised approaches which we can see by comparing Tables 7-11. Regarding topic modeling in section 6.2.2 we find that with LDA the accuracy score is significantly higher compared to without LDA especially across the three models. The net improvement due to LDA in accuracy is limited and the best F1 with LDA is lower than the best F1 without LDA. Further, we find that K-Means++ dominating the performance throughout. We argue the net improvement of F1 is lower in LDA because of the noisy data and number of topics so present in the dataset. Also our experiments confirmed that supervised models are still reliable for phrase-rubric relationship classification.

Finally in generic keyphrase-rubric relationship classification (section 6.3), we saw that traditional approaches like SVM with TF-IDF achieves consistently the highest performance across the folds, and this is the best traditional approach to generalizing from training to testing data (section 6.3.1). Also BERT trained with SVM as good predictor of F1 in specific keyphrase-rubric relationship classification, in which the BERT model was used as a feature extractor for each folds of the test corpus (section 6.3.2). This supported our hypothesis that language models without any training would still be useful owing to their training on large datasets, and training on a small dataset will negatively be correlated with F1. Also our experiments in section (section 6.3.3) confirmed the issue of reliability where traditional models are more reliable in terms of interpretability while language models without training are sensitive in nature. At the same time, we did see issue with RoBERTa where with SVM it produces the least result, which suggests need of more in-depth architectural exploration. Finally we also saw issue of mixed results for interpretability in section 6.3.3, we conjecture that this may be because of lack of training of the language models, again this needs to analyzed.

Also for we see that for all the three RQ's, by studying performance with the weighted average, it becomes clear that there is also a big difference in performance on corpora with the balanced and imbalanced dataset. This indicates that annotating more training examples for diverse classes would likely lead to a dramatic increase in F1 which in turn is expected to improve performance across all the explored RQ's.